\pdfoutput=1

\documentclass[11pt]{article}

\usepackage[final]{acl}

\usepackage{times}
\usepackage{latexsym}

\usepackage[T1]{fontenc}

\usepackage[utf8]{inputenc}

\usepackage{microtype}

\usepackage{inconsolata}

\usepackage{amsmath, amsthm, amssymb, amsfonts}
\usepackage{float}
\usepackage{multicol, multirow}
\usepackage{xcolor}
\usepackage{lipsum}
\usepackage{booktabs}
\usepackage{url}
\usepackage{graphicx}
\usepackage{adjustbox}

\usepackage[colorinlistoftodos,textsize=footnotesize]{todonotes}
\usepackage{amsmath}
\usepackage{amssymb}

\title{To Word Senses and Beyond: Inducing Concepts\\with Contextualized Language Models}

\author{  Bastien Li\'etard \and Pascal Denis \and
        Mikaela Keller \\
        University of Lille, Inria, CNRS, Centrale Lille,
        \\UMR 9189 - CRIStAL, F-59000 Lille, France \\ \texttt{first\_name.last\_name@inria.fr}}

\begin{document}
\maketitle
\begin{abstract}
Polysemy and synonymy are two crucial interrelated facets of lexical
ambiguity. While both phenomena are widely documented in lexical resources and have been studied extensively in NLP,
leading to dedicated systems, they are often being considered
independently in practictal problems. While many tasks dealing with polysemy (e.g. Word Sense
Disambiguiation or Induction) highlight the role of word's senses,
the study of synonymy is rooted in the study of concepts, i.e. meanings
shared across the lexicon. In this paper, we introduce Concept
Induction, the unsupervised task of learning a soft clustering among
words that defines a set of concepts directly from data.  This task
generalizes Word Sense Induction. We propose a bi-level
approach to Concept Induction that leverages both a local
lemma-centric view and a global cross-lexicon view to induce
concepts.  We evaluate the obtained clustering on SemCor's annotated
data and obtain good performance (BCubed F\textsubscript{1} above
0.60).  We find that the local and the global levels are mutually
beneficial to induce concepts and also senses in our setting. Finally,
we create static embeddings representing our induced concepts and use
them on the Word-in-Context task, obtaining competitive performance
with the State-of-the-Art.
\end{abstract}

\section{Introduction}
\label{sec:intro}

A crucial challenge in understanding natural language comes from the fact
that the mapping between word forms and lexical meanings is
many-to-many, due to polysemy (i.e., the multiplicity of meanings for
a given form)\footnote{In this paper, we take polysemy in its most
comprehensive definition, also including homonymy.} and synonymy (i.e., the multiplicity of forms for
expressing a given meaning). Both polysemy and synonymy have been
thoroughly studied in NLP, but mostly as independent problems, giving
rise to dedicated systems. Thus, Word Sense Disambiguiation (WSD) aims at
correctly mapping word occurrences to one of its
\textit{senses} \citep{raganato-etal-2017-word}, while Word Sense Induction
(WSI), its unsupervised counterpart, aims at clustering word
occurrences into latent senses directly from
data \citep{manandhar-etal-2010-semeval,
jurgens-klapaftis-2013-semeval}. More recently, researchers have
proposed the task of Word-in-Context (WiC), which consists in
classifying pairs of word occurrences depending on whether they
realize the same sense or
not \citep{pilehvar-camacho-collados-2019-wic}. All these works take a
word centric view, which aims at identifying or characterizing the
different senses of a given word, where these senses are bound to a
word. Another line of work, which takes a broader lexicon-wide
perspective, is concerned with identifying synonyms, which are
equivalence classes over different words that point to the same
concept \citep{zhang-etal-2021-self,ghanem-etal-2023-benchmark}, where \textit{concepts} are semantic entities that are not
bound to a word. In WordNet \citep{miller-1995-wordnet,fellbaum-1998-wordnet}, concepts are called \textit{synsets}, defined as sets of synonyms.
However, outside of lexical resources, synonymy and polysemy are usually considered as independent problems in the NLP literature.
Yet, these two views are complementary.
In lexicology, they correspond to two perspectives on the word-meaning mapping: \textit{semasiology} and \textit{onomasiology}.
The former is the word-to-meanings view, where one can observe polysemy by looking at the different meanings a given word has.
The latter is the meaning-to-words view, in which one can study synonymy by looking at the inventory of words that speakers use to express the same meaning.




In this paper, we propose a new task, called Concept Induction, that
directly aims at learning concepts in an unsupervised manner from raw
text. More precisely, this task aims at learning a soft clustering
over a target lexicon (i.e., a set of words), in such a way that each
cluster corresponds to a (latent) concept. Thus, this task both
addresses polysemy (since polysemous words should appear in multiple
clusters) and synonymy (since synonymous words should appear in the
same cluster(s)). 
Inducing concepts can be interesting for many external applications, like building lexical resources for low-resources languages \citep{velasco-etal-2023-towards}, and can bring a different perspective in computational studies of meaning, moving the usual word-centric focus to a more meaning-centric state. 

Our approach to Concept Induction relies on word occurrences for a
target lexicon, represented as word embeddings derived from a
Contextualized Language Model (in this case, BERT Large \citep{devlin-etal-2019-bert}), which are
then grouped, using hard clustering algorithms, into concept denoting
clusters. While these concept clusters could in principle be obtained
directly from word occurrences, we propose a bi-level methodology that
leverages both a \textit{local}, lemma-centric clustering (i.e.,
operating on only specific word occurrences), and a \textit{global},
cross-lexicon clustering (i.e., operating on all words
occurrences). From this perspective, our approach generalizes, and in
fact builds upon classical Word Sense Induction, in that word senses
are learned jointly alongside with concepts.
We hypothesize that an approach taking both complementary resolutions in account will lead to improved Concept Induction and Word Sense Induction, i.e. that the two objectives can be mutually beneficial.

To validate our approach, we carried out experiments on the SemCor
dataset, which provides a set of concepts (taking the form of
WordNet synsets) related to word occurrences. We found that our
bi-level clustering approach accurately learn concepts, achieving
$F_1$ scores above $0.60$ on the task of Concept Induction compared to
WordNet's synsets, outperforming competing approaches that use only
local and global views. This demonstrates the benefits of our bi-level
approach, and its ability to leverage both local and global views when
inducing concepts. Interestingly, we show that the benefits go both
ways: our proposed approach outperforms lemma-centric approaches when
evaluated for WSI. Finally, we show that concept-aware static
embeddings derived from our approach are also competitive with
state-of-the-art approaches efficient on the Word-in-Context task,
while using less training data. Through the new task of concept
induction, we also contribute in a new way to the ongoing debate
regarding the ability to align vector representations extracted from
Contextualized Language Models to the semantic representations posited
by (psycho-)linguists. In this vein, we conduct a qualitative
evaluation of obtained clusters to ensure they indeed reflect concepts
and gather synonyms. The source code we used for experiments is available at \url{https://github.com/blietard/concept-induction}.

\section{Related Work}
\label{sec:related}

\subsection{Lexical resources for concepts}

Princeton's WordNet (PWN) \citep{miller-1995-wordnet, fellbaum-1998-wordnet} is a lexical database that has been been the most widely used as a reference for most wordsense-related tasks for many years.
In WordNet, the entry corresponding to a lemma has different wordsenses, each of them mapping to a \textit{synset}.
Synsets are WordNet's equivalents of our concepts.
Lemmas whose wordsenses belong to the same synset are synonymous.
WordNet 3.0 contains 117,659 synsets and is built from the work of psycholinguists and lexicographers, that not only describes synonymy but also other lexical relations such as hypernymy/hyponymy, antonymy, meronymy/holonymy, etc.
But the amount of resources needed to create such lexical databases with human experts is considerable, making them a very rare and precious resource.
They are not available for a large number of active languages, and even more rare for dead languages \citep{bizzoni-etal-2014-making, khan-etal-2022-towards}.

\subsection{Word senses with Language Models}

With the recent development of neural Contextualized Language Models (CLM), several work use their hidden-layers to extract vector representations of word usages and retrieve word senses.
These representations are fed to a classification (for WSD) or a clustering (in the case of WSI) algorithm to distinguish the word's senses \citep{Scarlini-Pasini-Navigli-2020, nair2020contextualized, saidi2023sentence}.
These embeddings-based approaches have applications in other fields: \citet{kutuzov-giulianelli-2020-uio} and \citet{martinc-etal-2020-capturing} use sense clusters found using CLM embeddings to study the change in meaning of words, and \citet{chronis-erk-2020-bishop} propose a many-Kmeans method to investigate semantic similarity and relatedness.
Another line of work uses list of substitute tokens sampled from the CLM head to infer senses \citep{amrami-goldberg-2019-better, eyal-etal-2022-large} and are sucessful on WSI benchmarks like \citet{manandhar-etal-2010-semeval} and \citet{jurgens-klapaftis-2013-semeval}.

\subsection{Structures of Meaning in CLM}

Recent research probes neural CLMs for alignements between representations from their latent spaces and semantic patterns and relations.
Section 7.2 of \citet{haber-poesio-2024-polysemy} summarizes findings about polysemy in contextualized CLMs, showing that these models were able to detect polysemy and in some cases distinguish actual polysemy from homonymy.
They report that representations from different senses may however overlap.
\citet{hanna-marecek-2021-analyzing} shows that pretrained BERT embed knowledge of hypernymy but is limited to the more common hyponyms.

\citet{velasco-etal-2023-towards} build on top of WSI techniques in an attempt to automatically construct a WordNet for Filipino, thus proposing a modeling of synonymy in this language.
However, the evaluation of the synsets they obtained is limited by the lack of sense-annotated data for Filipino, and they could not evaluate the impact of their methodology on the two levels (senses and concepts).

Works like \citet{ethayarajh-2019-contextual} and  \citet{chronis-erk-2020-bishop} study the kind of information that was distributed across layers. 
The former concludes that syntactic and word-order information are distributed in the first layers while in deeper layers, representations are heavily influenced by contexts.
The latter demonstrates, with a multi-prototypes embedding approach, that semantic \textit{similarity} is best found in moderately late layers, while \textit{relatedness} is best found in last layers.

\newcommand{\card}[1]{\vert #1 \vert}
\newcommand{\vocab}{W}
\newcommand{\Corpus}{O}
\newcommand{\Occs}[1]{O^{#1}}
\newcommand{\occ}[2]{o^{#1}_{#2}}
\newcommand{\sensename}{s}
\newcommand{\sense}[2]{\sensename^{#1}_{#2}}
\newcommand{\Senses}[1]{S^{#1}}
\newcommand{\SSenses}{S}
\newcommand{\conceptname}{c}
\newcommand{\concept}[1]{\conceptname_{#1}}
\newcommand{\Concepts}{C}
\newcommand{\ConceptsLex}{\Concepts^{\vocab}}
\newcommand{\cluster}[1]{\hat{\conceptname}_{#1}}
\newcommand{\Clusters}{\hat{\Concepts}}
\newcommand{\ClustersLex}{\Clusters^{\vocab}}

\newcommand{\OccsToSenses}{f_{\Corpus\rightarrow\SSenses}}
\newcommand{\OTS}{\OccsToSenses}
\newcommand{\SensesToConcepts}{f_{\SSenses\rightarrow\Concepts}}
\newcommand{\STC}{\SensesToConcepts}

\section{Concept Induction}
\label{sec:conceptinduction}

Our main motivation behind Concept Induction is to present a view of the mapping between words and their meaning(s).\footnote{This mapping is called \textit{patterns of lexification} by \citet{françois-2022-lexical-tectonics}; see also \textit{coexpression} and \textit{synexpression} in the terminology proposed by \citet{haspelmath-2023-coexpression}.}
This view is systemic, meaning that it should not be defined for individual words neither for individual concepts, but rather acknowledging these as a whole with interactions and relations.
This extends beyond the primary objective of WSI, which defines word senses as pertaining to individual words only and does not explore relations between lemmas or concepts.

\subsection{Basic notions}
Consider a set of target words (or \textit{lemmas}) and for each lemma, we have a set of \textit{occurrences} of this word in a context (e.g. a sentence or a phrase).
The set of target lemmas is referred to as the \textit{lexicon}, while the \textit{corpus} is the set of all occurrences.
Our goal is to study the meaning of target words as they are used in the corpus.

In this study we call \textit{sense} of a word its usage to refer to a \textit{concept}.
A \textit{polysemous} word has multiple senses, each of them referring to a distinct concept.
Two words are said to be \textit{synonyms} for a given concept when each of them has one of their senses referring to this shared concept.
Senses are defined ``locally'', i.e. bound to an individual word of the lexicon, as opposed to concepts which are defined ``globally'', i.e. across the whole lexicon.
An occurrence of a word $w$ realizes one of $w$'s senses.
Consider the words ``test'' and ``trial'' and the following corpus:
(A) \textit{the jury found them guilty in a fair \underline{trial}}.
(B) \textit{candidates competed in a \underline{trial} of skill}.
(C) \textit{the hero underwent a \underline{test} of strength}.
The corpus is composed of two occurrences of ``trial'' and one occurrence of ``test.''
In the corpus, ``trial'' is polysemous. 
Its first sense, illustrated in A, refers to a \textit{process of law}.
Its second sense, in B, refers to the concept of \textit{the act of undergoing testing}.
The sense of ``test'' in sentence C also corresponds to this concept: it's a case where ``test'' and ``trials'' are synonymous.
Shifting the focus from senses to concepts, we will say that B and C instantiate the same concept, while A is an instance of a different concept.

\subsection{Task definition}

The goal of Concept Induction (CI) is to automatically learn a set of concepts directly from the data, i.e. learning a soft clustering $\ConceptsLex$ in the set of target words $\vocab$ that should correspond to the multiple concepts instantiated by occurrences of the corpus.
$\ConceptsLex$ is a \textit{soft} clustering because a word can be assigned to several clusters (when it is polysemous).
Using a different perspective than WSI, the framework of Concept Induction provides a more complete view on meaning across the lexicon.
Both WSI and CI capture \textit{polysemy}, but CI also reveals \textit{synonymy} across the lexicon.
Like WSI, Concept Induction does not require a pre-defined set of concepts.

\subsection{Formal framework}
\label{sec:ci:partitions}

\begin{figure*}[ht]
    \centering
    \begin{tikzpicture}[scale=1.3, every node/.style={scale=0.8}]
        \node (trial) at (0, 3) {``trial''};
        \node (test) at (0, 1) {``test''};
        \node[scale=1.4] (sd1) at (4, 3) {$\sense{trial}{1}$ };
        \node[scale=1.4] (sd2) at (4, 2) {$\sense{trial}{2}$ };
        \node[scale=1.4] (sb1) at (4, 1) {$\sense{test}{1}$ };
        \node[scale=1.4] (k1) at (6, 3) {$c_1$ };
        \node[scale=1.4] (k2) at (6, 1) {$c_2$ };
        \node[scale=1] (c1) at (8, 3) {$\{trial\}$ };
        \node[scale=1] (c2) at (8, 1) {$\{trial, test\}$ };
        \node[scale=1, align=center] (cd1) at (10, 3) {``process of law to\\ determine guilt.''};
        \node[scale=1, align=center] (cd2) at (10, 1) {``a challenge to\\ evaluate a skill'' };
        
        \node[scale=1] (w) at (0,-.5) {$\vocab$};
        \node[scale=1] (o) at (2,-.5) {$\Corpus$};
        \node[scale=1] (s) at (4,-.5) {$\SSenses$};
        \node[scale=1] (c) at (6,-.5) {$\Concepts$};
        \node[scale=1] (cw) at (8,-.5) {$\ConceptsLex$};

        \node[color=black!60] (forms) at (0,-1) {\textit{Words}};
        \node[color=black!60] (occs) at (2,-1) {\textit{Occurrences}};
        \node[color=black!60, align=center] (senses) at (4,-1) {\textit{Senses}};
        \node[color=black!60, align=center] (conceptsK) at (6,-1) {\textit{Concepts}\\\textit{(occurrence-level)}};
        \node[color=black!60, align=center] (concepts) at (8,-1) {\textit{Concepts}\\\textit{(Word-level)}};
        \node[color=black!60] (defs) at (10,-1) {\textit{Definitions}};

        \node (d11) at (2, 3.6) {$\occ{trial}{1}$};
        \node (d12) at (2, 3.2) {$\occ{trial}{2}$};
        \node (d21) at (2, 2.8) {$\occ{trial}{3}$};
        \node (d22) at (2, 2.4) {$\occ{trial}{4}$};

        \node (b11) at (2, 1.4) {$\occ{test}{1}$};
        \node (b12) at (2, 1.0) {$\occ{test}{2}$};
        \node (b21) at (2, 0.6) {$\occ{test}{3}$};

        \draw  (trial.east) to (d11);
        \draw  (trial.east) to (d12);
        \draw  (trial.east) to (d21);
        \draw (trial.east) to (d22);

        \draw  (test.east) to (b11);
        \draw  (test.east) to (b12);
        \draw  (test.east) to (b21);

        \draw [-] (d11) to (sd1.west);
        \draw [-] (d12) to (sd1.west);
        \draw [-] (d21) to (sd1.west);
        \draw [-] (d22) to (sd2.west);

        \draw [-] (b11) to (sb1.west);
        \draw [-] (b12) to (sb1.west);
        \draw [-] (b21) to (sb1.west);

        \draw [-] (sd1) to (k1);
        \draw [-] (sd2) to (k2);
        \draw [-] (sb1) to (k2);

        \draw [dotted] (k1) to (c1.west);
        \draw [dotted] (k2) to (c2.west);
    \end{tikzpicture}
    \caption{Illustration of our framework. The words ``trial'' is polysemous and has two senses corresponding to two different concepts, and is synonym with ``test'' for this second meaning.}
    \label{fig:framework}
\end{figure*}

Let $\vocab$ be the \textit{lexicon}.
For all word $w$ in $\vocab$, we denote $\occ{w}{i}$ the $i$-th occurrence of $w$ in the corpus.
We define $\Occs{w}~=\left\{\occ{w}{i}\right\}_{1\leq i \leq m_w}$ the set of $m_w$ occurrences of $w$.
The corpus, denoted $\Corpus$, is the union of all $\Occs{w}$.

For a given word $w\in\vocab$, the set $\Occs{w}$ can be partitionned according to its different senses.
We denote $\sense{w}{j}$ the part of occurrences of $w$ in the corpus corresponding to the $j$-th sense of $w$.
We refer to these groups of occurrences as the \textit{sense clusters} of $w$.
The set $\Senses{w}~=\{\sense{w}{j}\}_{1\leq j \leq n_w}$ forms a partition of
$\Occs{w}$, and we call $\SSenses$
the set of all sense clusters of all words,
i.e. $\SSenses=\bigcup_{w\in\vocab}\Senses{w}$.
$\SSenses$ is a ``local'' (lemma-centric) partition of the whole
$\Corpus$.
The task of Word Sense Induction aims at learning the partition $\SSenses$ given a corpus $\Corpus$.

In this work, we aim at dividing the corpus into \textit{concepts} instead of senses.
We denote $\concept{k}$ the group of occurrences of words corresponding to
the concept indexed by $k$, and $\Concepts~=\left\{\concept{k}\right\}_{1 \leq k \leq p}$ the partition of $\Corpus$ in $p$ concept clusters.
Unlike sense clusters of $\SSenses$, a concept cluster $\concept{k}\in\Concepts$ can gather occurrences of different words: $\Concepts$ is a ``global'' partition.
Each occurrence $\occ{w}{i}$ of a word $w\in\vocab$ is associated to a sense cluster $\sense{w}{j}$ and a concept cluster $\concept{k}\in\Concepts$.
We can say that a concept corresponding to $\concept{k}$ is instantiated by occurrence $\occ{w}{i}$ through the sense corresponding to $\sense{w}{j}$, or conversely that $\occ{w}{i}$ uses the sense reflected in $\sense{w}{j}$ to mean the concept described by concept cluster $\concept{k}$.
All occurrences of sense cluster $\sense{w}{j}\in\SSenses$ appear in the same concept cluster $\concept{k}\in\Concepts$.

In summary, $\SSenses$ and $\Concepts$ are partitions of $\Corpus$ and are naturally constrained as follows:
\vspace{-5pt}
\begin{enumerate}\itemsep=-2pt
    \item\label{cons:sense1word} By definition, a sense in $\SSenses$ is associated to one and only one word $w\in\vocab$.
    \item\label{cons:hardS} An occurrence $\occ{w}{i}$ realizes exactly one sense $\sense{w}{j}\in\SSenses$
    \item\label{cons:hardC} An occurrence $\occ{w}{i}$ instantiates exactly one concept $\concept{k}\in\Concepts$.
    \item\label{cons:unicity} In a given sense $\sense{w}{j}\in\SSenses$, all occurrences are assigned to the same concept $\concept{k}\in\Concepts$.
    \item\label{cons:distinct} All $\sense{w}{j}\in\Senses{w}$ (i.e. same word) refer to \textit{distinct} concepts.
\end{enumerate}
\vspace{-5pt}

From the partition $\Concepts$ on occurrences, one can derive $\ConceptsLex$, a clustering of the set of words $\vocab$ into concepts.
To each concept cluster $\concept{k}\in\Concepts$ we associate a cluster in $\ConceptsLex$ that contains all lemmas of $\vocab$ whose occurrences were assigned to $\concept{k}$.
In $\ConceptsLex$, a polysemous word with $n$ senses appears in $n$ distinct clusters (one per sense), and synonyms appear in at least one common cluster (one per shared concept).

We denote $\ClustersLex$ the word-level soft-clustering and $\Clusters$ the partition of occurrences that are \textit{learned} on the data.

In Figure \ref{fig:framework} we illustrate this framework, using a corpus of occurrences of the words ``test'' and ``trial''. 
In this scenario, $\vocab=\{\text{``test'', ``trial''}\}$ and two concepts are instantiated: \textit{a process of law to determine someone's guilt} and \textit{a challenge to evaluate a skill}.
The lemma ``trial'' exhibits two senses as it has occurrences corresponding to both concepts: ``trial'' is polysemous.
The second concept is also instantiated by occurrences of ``test'', therefore ``trial'' and ``test'' show synonymy in this case.
This toy example also follows all constraints formulated above.

\newcommand{\Csense}[2]{\hat{\sensename}^{#1}_{#2}}
\newcommand{\CSenses}[1]{\hat{\Senses{#1}}}
\newcommand{\CSSenses}{\hat{\SSenses}}

\section{Methodology}
\label{sec:methods}

In this section we describe the methods we propose and evaluate for Concept Induction.
We learn a clustering $\ClustersLex$ drawing inspiration from the relations between $\Corpus$, $\SSenses$, $\Concepts$ and $\ConceptsLex$. In particular, the overall objective of our methodology consist in finding $\Concepts$ (i.e. partition occurrences into concept clusters) to derive $\ConceptsLex$.
Section \ref{sec:ci:partitions} highlighted that there are two levels of partitions: a local level (senses) and a global one (concepts).
The proposed approaches rely on both levels and the use of a Contextualized Language Model (CLM) to gather representations of occurrences influenced by the context.

\subsection{Proposed Bi-level Method}

\paragraph{Local (lemma-centric) clustering}
Firstly, we propose to learn a word-sense partition for each target words individually.
Using the CLM hidden layers, we extract a vector representation (the occurrence embedding) of every occurrence $\occ{w}{i}$. 
We then learn a partition $\CSenses{w}$ of each $\Occs{w}$ using a clustering algorithm on the embeddings.
Each $\CSenses{w}$ describes the \textit{locally estimated} sense clusters of word $w$.
Jointly considering these partitions for all $w\in\vocab$, we obtain a partition $\CSSenses$ of the whole set of occurrences $\Corpus$.
This partition is \textit{local} in the sense that each word has its occurrences clustered independently from other words.

\paragraph{Global (cross-lexicon) clustering}
Once we have a local clustering $\CSSenses$, we turn from considering words independently to consider all words together.
In this step, we learn a \textit{global} clustering by merging local clusters of occurrences.
To do so, we average embeddings of all occurrences in the same local cluster to get a single embedding representing each local cluster.
Then we run a second clustering algorithm, this time using the averaged embeddings of local clusters.
This global clustering defines a new partition $\Clusters$ of the the corpus $\Corpus$:
when two local clusters $\Csense{w_1}{j}$ and $\Csense{w_2}{j^\prime}$ are merged into the same global cluster $\cluster{k}$ (because their embeddings were clustered together), all their occurrences are assigned to global cluster $\cluster{k}$.
From this global occurrence partition $\Clusters$ we can easily extract $\ClustersLex$, a word-level soft-clustering of lemmas whose occurrences appear in the same $\cluster{k}$.



This \underline{Bi-level} method directly implements the system of contraints described in Section \ref{sec:ci:partitions}.
Only constraint \ref{cons:distinct} is not enforced by design. 
Indeed, our local clusters being learned and not informed by an expert, the local clustering step may make errors, especially if the data for a given word are sparse.
Allowing the global clustering to merge local clusters enables the correction of local clustering's recall errors using information from the global level.

We also want to highlight that the proposed methodology is generic, in the sense that it is not tied to a specific choice of clustering algorithm.

\subsection{Local-only and Global-only}

Sense-inducing systems (WSI approaches) that create only local clusters of occurrences for each word are said to be \underline{Local-only} systems.
We use them as baseline models that only produce word-level clusters of size 1 and do not reflect synonymy, but still learn polysemy.

On the other hand, consider a system in which each occurrence is mapped to its own local cluster (i.e. no actual local clustering step), and the global step divides occurrences directly into global clusters.
We refer to this kind of system as \underline{Global-only} approaches.
They allow to evaluate how useful the local clustering step is in the process: we hypothesize that the local step in Bi-level will reduce potential variance in occurrences by aggregating them, increasing Precision compared to Global-only.

\newcommand{\fscore}{F\textsubscript{1}}
\newcommand{\CSim}{\operatorname{CSim}}
\newcommand{\MergedSynsets}[1]{\sigma_{#1}}



\section{Experiments}

In this section, we evaluate the abilities of the proposed methods to induce concepts and compare the proposed bi-level approach to other methods.
We investigate the advantages of the bi-level approach not only for the global viewpoint but also in the local setting.

\subsection{Settings}

\paragraph{Data.}
We choose to use the annotated part of the \textit{SemCor 3.0}\footnote{\url{http://web.eecs.umich.edu/~mihalcea/downloads.html\#semcor}} corpus.
This dataset contains occurrences for a wide number of words, and morpho-syntactic annotations provide their lemma and their Part-of-Speech tag.
Among all lemmas having at least 10 annotated occurrences, we keep only nouns (excluding proper nouns)\footnote{For the sake of simplicity and clarity, this study is focused only on nouns. Indeed, other Parts-of-Speech induce extra difficulties. Verbs for instance required extra preprocessing steps and decisions (e.g. include or exclude gerundive uses, past participle employed like adjectives, etc.). Extension of experiments to other PoS is left to future work.} composed only of alphabetical characters with a minimum length or 3 letters.
The resulting lexicon $\vocab$ contains 1,560 different lemmas, for which we gather a corpus $\Corpus$ containing a total of 52,997 occurrences\footnote{Sentences in which the lemma appears, paired with its position within them. If the lemma appears multiple times in the same sentence, we create several distinct occurrences, where only the position varies.}.
SemCor is also semantically annotated, with each occurrence of a target lemma assigned to a synset in WordNet, that we consider to be the concept it refers to.
We derive a reference partition of the occurrences $\Concepts$ and a reference soft-clustering of the words $\ConceptsLex$ from annotations, for a total of 3,855 different concepts (WordNet's synsets) covered in $\Corpus$. This set of concepts is the subset of WordNet corresponding to the textual data.

\paragraph{Evaluation of Concept Induction}
We compare the learned word clustering $\ClustersLex$ to the reference $\ConceptsLex$.
We choose to use the BCubed metrics \citep{bagga-baldwin-1998-entity-based}, obtaining Precision and Recall for the evaluated clustering compared to the reference, as well as an $F_1$ score.
To account for overlapping clusters, we use the Extended BCubed metrics proposed by \citet{amigo-etal-2009-comparison}, which has already been used as evaluation in SemEval2013 WSI task \citep{jurgens-klapaftis-2013-semeval}.

Using BCubed metrics, for a given evaluated clustering,
low precision would mean that grouped lemmas should not have been clustered together because none of their occurrence annotations map to a shared concept according to annotations .
Low recall means that the evaluated system fails to capture clusters of lemmas whose occurrences share a concept according to annotations.
The number of common clusters between two words also impacts BCubed metrics: if two lemmas appear together in too many clusters compared to the reference clustering, precision is decreased; if the number of common clusters is too low, recall is decreased.

\paragraph{Development.}
To learn the clustering, candidate systems have access to the full set of occurrences-in-context but not their annotations.
To choose the appropriate set of hyperparameters, we create a \textit{Dev} split of the annotations by randomly sampling 10\% of concepts and revealing semantic annotations of the corresponding occurrences.
We use them to evaluate Concept Induction for this small set of concepts, and choose the set of hyperparameters that scores best in BCubed $F_1$.

\paragraph{Evaluation splits}
In the final evaluation phase, we compute scores on all concepts / all occurrences, including the \textit{Dev} split, as concepts in it are part of the whole subset of WordNet described by SemCor's annotations.
In the full data, we found that 88\% of the concepts were instantiated using only a single lemma.
To better evaluate cases of synonymy, we also evaluate systems on a subset of the corpus, denoted ``\textit{Synon}'', that contains only occurrences of concepts showing synonymy (the remaining 12\% of concepts, instantiated through at least 2 distinct lemmas).
Statistics are provided in Table \ref{tab:corpus} in Appendix \ref{sec:appdx:splits}.
Note that it only changes the set of concepts/lemmas for which the system is being evaluated, not the clustering's training data.

\subsection{Systems and baselines}

\paragraph{Clustering Algorithms.}
We try two different clustering algorithms relying on different paradigms: Kmeans (used in \citet{chronis-erk-2020-bishop}), a centroid-based algorithm with a fixed number of clusters, and Agglomerative clustering (used in \citet{saidi2023sentence,velasco-etal-2023-towards}; dubbed ``Agglo'' for short), a deterministic hierarchical approach using a distance threshold to create a dynamic number of clusters instead of using a fixed one.
Another difference between Kmeans and Agglo is that the former assumes that expected clusters are of nearly-spherical shape and balanced in number of points, while the latter does not make assumptions on the shape of data.
Details of tested hyperaprameter values are provided in Appendix \ref{sec:appdx:hyperparams-layers}.

\paragraph{Representations.}
Following \citet{chronis-erk-2020-bishop} and \citet{eyal-etal-2022-large}, we use BERT Large \citep{devlin-etal-2019-bert}, a \textit{masked} language model with 24 layers and 345M parameters. 
This allows for direct comparisons with these approaches. 
Also, BERT Large was found by \citet{haber-poesio-2021-patterns-polysemy} to allow for better grouping of sense interpretations than other LLMs.\footnote{We leave to further work the use of autoregressive and/or newer Language Models.}
We average subwords' embeddings if needed.
It is a common practice in previous work on semantic-related tasks to use the average of the last 4 layers to get embeddings; we decided to adopt the same "4 layers average pooling" strategy, but trying with different possible sets of layers (see Appendix \ref{sec:appdx:hyperparams-layers}).
Therefore, for a set of four layers, we average hidden states across the selected layers to get a single 1024-dimensional vector.
We found that layers 14 to 17 obtained the best results on \textit{Dev} for all methods (global/local-only and bilevel).

\paragraph{Sense-inducing systems.}
Comparison to Local-only systems will give a (strong) baseline just by inducing senses without aiming at concepts.
We used the same clustering algorithms.
We also implement the WSI method proposed by \citet{eyal-etal-2022-large}.
It relies on a different paradigm, using the Language Model for \textit{substitution} instead of word embeddings.
From lists of substitutes, they build a graph of substitutes in which they find communities and then assign each occurrence to a community of substitutes to find the wordsenses.
Because Local-only methods only induce senses, their hyperparameters are chosen to maximize a WSI objective on polysemous words of the dev split.

\paragraph{Baselines}

We construct a candidate clustering $\ClustersLex$ where each lemma has its own cluster. 
This baseline model is referred to as the ``Lemmas'' baseline.
This is to evaluate the extent to which the information contained by the lemma alone can be used to induce concepts without any knowledge on word senses neither on context.
As a second baseline, we create for each lemma as many singletons as the number of different concepts its occurrences are annotated with.
All created clusters are of size 1: we account perfectly for polysemy but not at all for synonymy.
This second baseline is dubbed ``Oracle WSI''.

\subsection{Concept Induction in SemCor}

\begin{table}[t]
    \adjustbox{max width=\columnwidth}{%
    \small
    \centering
    \begin{tabular}{lccc|ccc}\toprule
        & \multicolumn{3}{c}{\textbf{Full data}} & \multicolumn{3}{c}{\textbf{Synon.}} \\
        \textbf{Concept Induction} & P & R & \fscore & P & R & \fscore \\\midrule
        \multicolumn{1}{c}{\textbf{Baselines}} & & & & \\
        Lemmas & 1.0& .43 & .61 & 1.0 & .61 & .50 \\
        Oracle WSI & 1.0 & .75& .86 & 1.0& .39 &.56 \\
        \multicolumn{1}{c}{\textbf{Local-only Systems}} & & & & \\
        Kmeans Local & .73 & .70 & .71 & .67& .38& .49\\
        Agglo Local & .92 & .53 & .67 & .92 & .35& .50\\
        \citet{eyal-etal-2022-large} & .31& .75& .44 & .37& .39& .38\\
        \multicolumn{1}{c}{\textbf{CI Systems}} & & & & \\
        Kmeans Global & .48 & .65 & .56 & .68 & .54 & .60\\
        Kmeans Bi-level & .70 & .59 & .64 & .82 & .47 & .59\\
        Agglo Global & .61 & .60 & .60 & .82 & .50 & .62\\
        Agglo Bi-level & .75 & .60 & .66 & .86 & .49 & .62\\
        \bottomrule
    \end{tabular}
    }
    \caption{Concept Induction BCubed Precision (P), Recall (R) and \fscore on the SemCor data averaged over 5 runs.}
    \label{tab:ci-overall}
    \vspace*{-1em}
\end{table}

In Table \ref{tab:ci-overall} we display the Concept Induction scores (\fscore) of proposed baselines and systems on the full SemCor data and on the \textit{Synon.} split.
On the full data, both the Lemmas and Oracle WSI baselines achieve very good performance because they have, by design, a perfect precision (they do not cluster lemmas at all and do not overestimate the number of clusters) and because 88\% of concepts are instantiated with only a single lemma (thus their recall is still good).
However, they are very limited on the \textit{Synon.} split of the data, where concepts are instantiated with multiple lemmas.

The proposed Concept Induction systems reach scores ranging from .56 to .66 on the full data, half of them outperforming the Lemmas baseline, and from .59 to .62 on the \textit{Synon.} split, outperforming all other systems.
While still challenging, it exhibits that it is indeed possible to induce WordNet-based \textit{concepts} in a corpus using LMs hidden layers vectors.

We also see that Kmeans-based approaches are consistently outperformed by Agglomerative methods.
This indicates that the representational spaces in LM hidden layers are not organized in a nearly-spherical fashion as Kmeans algorithm assumes, but rather are populated less uniformly.
This is reflected in precision and recall: Agglomerative systems reach a higher precision than Kmeans with similar recall.

Overall, results are in favor of \textit{Bi-level} approaches over \textit{Global-only} systems, with substantial improvements in \fscore\, on the full data while obtaining (nearly) identical performance on concepts of multiple lemmas, and large increases in precision while the loss in recall is minimal.
This demonstrates that considering the local (lemma-centric) perspective is beneficial to a global (cross-lexicon) view when inducing concepts.
The local clustering, with the subsequent representation averaging, helps reducing variance in occurrences and therefore allow to reach higher levels of precision in the global clustering compared to Global-only.
We would also like to emphasize that, while {Global-only} systems are more simple in design, their computational cost is usually higher than {Bi-level} ones, especially when the clustering algorithm's time complexity is quadratic with respect to the number of occurrences.

\subsection{Qualitative Analysis of Concepts Clusters}

\begin{table}
    \small
    \centering
    \begin{tabular}{lccc}
        \toprule
        & \multicolumn{3}{c}{\textbf{Cluster size}}\\
         & 2 & 3 & 4+ \\
        \midrule
        \textbf{Nb. of annotated clusters} & 50 & 50 & 23\\
        \midrule
        \textbf{Category (\% of annotated clusters)} &  &  & \\
        Synonyms & 42 & 38 & 17 \\
        Near-synonyms & 24 & 24 & 35 \\
        Related & 26 & 36 & 48 \\
        Invalid & 08 & 02 & 0 \\
        \bottomrule
    \end{tabular}
    \caption{Qualitative manual evaluation of obtained word clusters of size $\geq 2$.}
    \label{tab:quali}
    \vspace*{-1em}
\end{table}

We manually annotate word clusters (obtained from our best-performing approach, the Agglo Bi-level system) containing at least 2 lemmas according to the semantic similarity between lemmas.
Distribution of cluster sizes (in number of lemmas) can be found in Appendix \ref{sec:appdx:size-distrib}.
We distinguish four categories:
\textit{synonyms} when lemmas are cognitive synonyms (e.g. ``necessity'' and ``need''), \textit{near-synonyms} for lemmas close to be synonyms but showing slight difference in meaning (e.g. ``duty'' and ``task'', the former being stronger than the latter),\footnote{Notions of \textit{cognitive synonymy} and \textit{near-synonymy} are discussed by \citet{stanojevic-2009-cognitive}.} \textit{related} when lemmas show a topical (e.g. ``dirt'', ``sand'' and ``mud'') or lexical relations (e.g. antonyms like ``man'' and ``woman'') and \textit{invalid} clusters when lemmas show no semantic relation (e.g. ``child'' and ``idea'').

Proportions of these annotations are displayed in Table \ref{tab:quali} with respect to the cluster size, the number of lemmas in the cluster.
For a given cluster size, if the number of clusters exceeds 50, we randomly sample 50 clusters to be annotated.
Overall, the proportion of synonyms and near-synonyms is generally above 50\% and less than 10\% of clusters are invalid, indicating that most learned concepts are reliable and meaningful.
We argue that the remaining \textit{related} term clusters, while not synonyms, may still be interesting in less fine-grained studies.
The portion of \textit{related} clusters is in line with findings from previous work showing that BERT was also reflective of other lexical relations, such as hypernymy \citep{hanna-marecek-2021-analyzing}.

\subsection{Benefits at the Local Level}

We now turn back to the local level and assess whether the information brought at the global level helps distinguishing senses of individual words.
Here we do not evaluate the word-level soft clustering, but the occurrence-level division of SemCor's data, considering each word independently.
In other words, we evaluate WSI in SemCor using annotations as the reference sense clustering.

\paragraph{Evaluation of induced senses}
For each word $w\in\vocab$, we compare how its set of occurrences $\Occs{w}$ is divided in $\Clusters$ to how it is divided in the reference $\Concepts$ provided by annotations using BCubed metrics, and we average scores obtained across $\vocab$.
We display the WSI BCubed \fscore, as in previous WSI tasks like \citet{jurgens-klapaftis-2013-semeval}.
Following \citet{amrami-goldberg-2019-better}, we report $\rho$ the Spearman correlation coefficient between the number of clusters a lemma is assigned to and its number of senses according to annotations, to ensure that the number of created senses actually scales with the actual degree of polysemy.

Note that, for CI systems, we evaluate the division of occurrences provided by the \textit{final} clustering $\Clusters$ (i.e. how occurrences are clustered after the global step and its potential merge operations).
The quality of sense clusters induced by the local-step only is actually evaluated with Local-only systems.

\begin{table}
    \small
    \centering
    \begin{tabular}{lcc}\toprule
        & \textbf{WSI \fscore} & \textbf{$\mathbf{\rho}$}\\\midrule
        \multicolumn{1}{c}{\textbf{Local-only Systems}} & & \\
        Kmeans Local & .61 & NA \\
        Agglo Local & .77 &  .04\\
        \citet{eyal-etal-2022-large} & .46 & .51 \\
        \multicolumn{1}{c}{\textbf{CI Systems}} & & \\
        Kmeans Global & .76 & .51\\
        Kmeans Bi-level & .78 & .30 \\
        Agglo Global & .80 & .53\\
        Agglo Bi-level & .80 & .46 \\
        \bottomrule
    \end{tabular}
    \caption{WSI BCubed \fscore and sense number correlation coefficient $\rho$ on SemCor full data. Not computed for Kmeans because the number of cluster is constant.}
    \label{tab:wsi-overall}
\end{table}

\paragraph{Local results.} 
Results of this local evaluation are displayed in Table \ref{tab:wsi-overall}.
Let us recall that Local-only systems' hyperparameters are chosen to maximize the WSI \fscore on the dev split, while those of CI systems maximize the Concept Induction \fscore.
Nonetheless, one can observe that all CI systems outperform their Local-only counterparts, achieving higher WSI \fscore and $\rho$ even though their hyperparameters are not chosen to match the WSI itself.
This indicates that the information brought at the global level by considering cross-lexicon relations may indeed help improving WSI, and benefits between local and global levels go both ways.

We explain the relatively poor performance of State-of-the-Art WSI system by the fact that we are in a particular setting, where the number of occurrences per lemma is relatively low in SemCor (30 per lemma on average) and so is the average number of occurrences per concept.
Data sparsity is a favorable ground for word senses to be misrepresented.
As such, methods meant to be applied on larger datasets like the one of \citet{eyal-etal-2022-large} may not work as well as expected.
Our results show the limitations of these systems when the amount of training data is low and the interest of aiming at concepts to get senses.
This scenario is motivated in areas where data are not available in large quantities and still require to induce senses.
In the case of the study of Lexical Semantic Change (the evolution of word meanings over time), recent works perform WSI in diachronic corpora that are often unbalanced and small \citep{tahmasebi-etal-2021-survey}.


\section{Extrinsic Evaluation with Concept-aware Embeddings}

\begin{table}
    \small
    \centering
    \begin{tabular}{l|c}\toprule
            \textbf{Model} & \textbf{Acc.} \\\midrule 
            \citet{eyal-etal-2022-large} (CBOW) & 59.3 \\
            \citet{eyal-etal-2022-large} (Skip-Grams) & \textbf{61.9} \\\midrule
            Ours (Agglo global) & 58.8\\
            Ours (Agglo bi-level) & 59.7 \\\bottomrule
    \end{tabular}
    \caption{Accuracy scores on the nouns of the WiC test dataset \citep{pilehvar-camacho-collados-2019-wic}.}
    \label{tab:wic}
    \vspace*{-1em}
\end{table}

In their work, \citet{eyal-etal-2022-large} derive \textit{sense-aware} static embeddings from their WSI method, training them on the \textit{Wikipedia} dataset and used them for the Word-in-Context (WiC) task.
They achieve nearly-SotA results on the dataset proposed by \citet{pilehvar-camacho-collados-2019-wic}, and report to be outperformed only by methods using external lexical knowledge and resources.
We proceed to the same extrinsic evaluation of our work, constructing \textit{concept-aware} embeddings using concept clusters of Concept Induction systems (Global-only and Bi-level Agglo).
To obtain such embeddings, we average all vectors representating occurrences in SemCor contained each global cluster to get one vector per concept cluster.

The WiC task consists of determining whether two occurrences of a target lemma $w$ correspond to the same sense.
The WiC dataset's target words are nouns and verbs, but like in the rest of this paper, we restrict our scope to nouns.

To solve the task, we use BERT Large to create representations of the two target occurrences.
Each of them is assigned to a concept by finding the closest concept-aware using cosine distance.
The decision depends on whether the two occurrences are mapped to the same concept (\textit{true}) or to distinct ones (\textit{false}).
Results are displayed in Table \ref{tab:wic}.
Our concept-aware embeddings obtain very similar results to those of their sense-aware embeddings, with ours derived from our bi-level approach even outperforming their CBOW method.
Interestingly, our embeddings were trained with far fewer resources than theirs, as we used 52 997 occurrences from the SemCor dataset while they used a dump of Wikipedia, gathering millions of occurrences.
This emphasizes the value of concept-aware embeddings: the use of cross-lexicon information allows competitive results with fewer resources.

\section{Conclusion}
In this paper, we argued that, while word senses allow to investigate polysemy, concepts are a larger perspective that allows the study of polysemy as well as synonymy.
We defined Concept Induction, the unsupervised task to learn a soft-clustering of words in a large lexicon, directly from their in-context occurrences in a corpus.
Then, we proposed a formulation of this problem in terms of \textit{local} (lemma-centric) and \textit{global} (cross-lexicon) complementary views, and tested an approach that uses information from both levels using contextualized Language Models.
On concept-annotated SemCor corpus, we found that this bi-level view was beneficial for Concept Induction, and even for Word Sense Induction with a low amount of training data.
We validated the quality of obtained clusters with manual annotations, ensuring that clusters mostly correspond to actual synonyms and concepts.
Finally, we showcased an external application of our methodology to create concept-aware embeddings that can be competitive to other methods on semantic tasks, such as Word-in-Context.

Concept Induction opens the way for a different perspective on lexical semantics in NLP, and can be a basis for many studies of lexical meanings as it is expressive enough to reflect relations on both sides of the word-meaning mapping.

\section{Limitations}

The formal framework we defined uses terminology and notions from rather structuralist/relational assumptions of the language's lexical system (e.g. senses, discrete concepts, etc.).
We made this choice based on how lexical databases like WordNet (and its derivatives), or other like the Historical Thesaurus of English for instance, are designed using the "word/sense/concept" structure.
From a purely practical point of view, this choice makes sense as these resources would be the primary source for task data's annotations.
Conceptually, senses are also a notion widely used in computational linguistics and we wanted to propose Concept Induction as a step "beyond" this conventional aspect and its related tasks.
Future research may explore definitions/extensions of Concept Induction outside of this structuralist/relational framework, towards cognitive semantics for instance \citep{geeraerts-2010-theories}.

Evaluating Concept Induction is mainly limited by the low amount of suitable annotated corpora.
Not only the data need to be annotated in concepts, but these annotations must cover a wide variety of lemmas for synonymy to be sufficiently represented in the corpus.
Future work may find or create datasets meeting these requirements to evaluate Concept Induction outside of SemCor.

For now, the study is limited to nouns. Performances of benchmarked algorithms and systems may change with other Part-of-Speech tags.

Our Bi-level method allows the global clustering to \textit{merge} local clusters, leveraging lexicon-level information to be used to correct Word Sense Induction errors at the lemma-level.
By its sequential nature, our method does not allow to \textit{split} local clusters using global-level information, which could lead to better results.
Further research directions include creating an iterative version of our methodology (alternating local and global clustering), or attempting to tackle both clustering objectives simultaneously with bi-level constrained clustering.

Our results about sense-induction at the local level showed that usual WSI methods may not be robust in our setting where there are few occurrences for some lemmas.
We demonstrated that, in this setting, concept-inducing methods provided a better division in word senses.
In many fields of linguistics, corpora are not very large and do not contain hundreds of occurrences for each word.
Nonetheless, it is still uncertain if this observed advantage of CI systems would still hold on bigger datasets with many occurrences per lemma, a setting better-suited for usual WSI methods.

In this paper, we limited our study to Nouns, the morpho-syntactic class exhibiting the most prominent semantic features.
We leave to further research the study of Concept Induction for Verbs, Adjectives, or the heterogeneous family of Adverbs.

\section{Ethical Considerations}

Our methodology uses pretrained Contextualized Language Models, which are know to encode and replicate social biases contained in their training data and sometimes amplify them.
While we do not observe surface-level biases arising when manually annotating concept clusters, it is still an open question of how these social biases may influence or even change results when inducing concepts in SemCor.

\section*{Acknowledgements}

We gratefully thank the anynymous reviewers for their insightful comments.
This research was funded by Inria Exploratory Action COMANCHE.

\bibliography{anthology,custom}

\appendix

\section{Extended BCubed to Evaluate CI and WSI}
\label{sec:appdx:bcubed}

The extension of BCubed for overlapping clusters rely on two quantities, Multiplicity Precision (MP) and Multiplicity Recall (MR).
In the case of Concept Induction, MP and MR between two lemmas are defined as follows:
        \begin{multline*}
                \operatorname{MP}(w_1,w_2) = \\\frac{ Min\left(\vert f(w_1) \cap f(w_2) \vert , \vert g(w_1) \cap g(w_2) \vert \right)}{\vert f(w_1) \cap f(w_2) \vert}  \\                
        \end{multline*}
        \begin{multline*}
                \operatorname{MR}(w_1,w_2) = \\\frac{ Min\left(\vert f(w_1) \cap f(w_2) \vert , \vert g(w_1) \cap g(w_2) \vert \right)}{\vert g(w_1) \cap g(w_2) \vert}
        \end{multline*}
with $w_1$ and $w_2$ two lemmas, and $g$ a reference clustering function and $f$ the clustering function we want to evaluate.
MP (resp. MR) can be computed for every lemma $w_1$ with every other lemma $w_2$ sharing at least one cluster with $w_1$ in $f$ (resp. in $g$). We denote $\operatorname{MP}(w_1,\cdot)$ and $\operatorname{MP}(w_1,\cdot)$ the obtained averages.
In the case of non-overlapping clusters, this formulation gives the same result as the original (non-extended) BCubed.
To evaluate WSI, the formulation is the same but we do not evaluate at the word-level but at the occurrence-level.

Precision, Recall and F-score are obtained as follows:
\begin{gather*}
        \operatorname{Precision} = \frac{1}{\card{\vocab}}\sum_{w \in \vocab} \operatorname{MP}(w,\cdot)
        \\
        \operatorname{Recall} = \frac{1}{\card{\vocab}}\sum_{w \in \vocab} \operatorname{MR}(w,\cdot)
        \\
        \operatorname{F_\beta} = (1+\beta^2) \frac{\operatorname{Recall} \times \operatorname{Precision}}{\beta^2\times \operatorname{Precision} + \operatorname{Recall} }. 
\end{gather*}
By default we fix $\beta=1$, as we compare the learned clustering and the reference clustering as equals and therefore do not find that Precision and Recall should be weighted differently.

\citet{amigo-etal-2009-comparison} showed that the benefits of BCubed over other clustering scores.
For instance, Rand Index does not handle well the case of many small clusters, which is likely to be the case for Concept Induction.
We also prefer Extended BCubed over Overlapping Normalized Mutual Information \citep{mcdaid-etal-2011-normalized} as the latter is matching-based.
That is, the repetition (or non-repetition) of identical clusters will have no impact on the measure.
However, we can easily imagine identical clusters of words to be repeated as they may correspond to distinct concepts.
In Extended BCubed, repeated clusters are taken in account as we measure the number of times two lemmas are clustered together.
The denominator of MP ensures that over-estimating the number of common clusters is also penalized, and those of MR ensures that under-estimating is penalized.
$Min$ operators are there to prevent both quantities to grow over 1.

\section{Splits and dataset statistics}
\label{sec:appdx:splits}

In Table \ref{tab:corpus} we display statistics over the different splits we used.
Dev is a subset containing a sample of 10\% of concepts and their occurrences.
Synon. is a subset containing only concepts instantiated with 2 lemmas or more, and their occurrences. 

\begin{table*}[h]
        \small
        \centering
        \begin{tabular}{lrrr|rr|rr}
                \toprule
                & \#Occs & \#Lemmas & \#Concepts & \#Occs/Concept & \#Occs/Lemmas & $d_\text{Lex}$ & $d_\text{Polysemy}$ \\
                \midrule
                Full data & 52'997 & 1'560 & 3'855 & 13.75 & 33.97 & 1.14 & 2.83 \\
                Dev & 4795 & 389 & 386 & 12.42 & 12.33 & 1.14 & 1.13\\
                \textit{Synon} & 13'158 & 630 & 447 & 29.44 & 20.89 & 2.24 & 1.59 \\
                \bottomrule
        \end{tabular}
        \caption{Statistics on the different data splits in annotated SemCor. 
                The split ``Synon'' only contains occurrences of concepts instantiated with multiple lemmas (cases of synonymy).
                $d_\text{Lex}$ is the average number of unique lemmas per concept, $d_\text{Polysemy}$ is the average number of distinct concepts per lemma.
                }
        \label{tab:corpus}
    \end{table*}
    
\section{Used hyperparameters and layers}
\label{sec:appdx:hyperparams-layers}
\begin{table*}[ht]
        \small
        \centering

        \begin{tabular}{ll}\toprule
                \textbf{Systems} & \textbf{Best hyperparameters}\\\midrule
                Local-only Kmeans & $k=3$ \\
                Local-only Agglo & linkage = average, $\nu=1.0$\\
                Global-only Kmeans & $\pi=120\%$ \\
                Global-only Agglo & linkage = average,$\nu=3.5$\\
                Bi-level Kmeans & $k=8, \pi=120\%$\\
                Bi-level Agglo & linkage = average (both), $\nu_{local}=0.0, \nu_{global}=4.5$\\
                Bi-level Kmeans (local Agglo)& linkage = average $\nu_{local}=0.0, \pi=120\% $\\
                Bi-level Agglo (local Kmeans)& $k=10$, linkage = average, $\nu_{global}=4.5$ \\
                \bottomrule
        \end{tabular}
        \caption{Best hyperparameters on the \textit{Dev} split.}
        \label{tab:best-hypers}
\end{table*}

\subsection{CLM layers}
Prior work like \citet{ethayarajh-2019-contextual} showed that later layers usually correlates with deeper levels of contextualization and more semantic information, \citet{chronis-erk-2020-bishop} showed that moderately-late were preferred for lexical similarity while very last layers were preferred for semantic relatedness.
To get embeddings, we try 4 sets of layers corresponding to different depths: first layers (1 to 4), moderately early layers (8 to 11), moderately late (14 to 17), and last layers (21 to 24).
To get the representation of a word's occurrence, we simply average its embeddings from the four chosen layers into one single 1024-dimensional embedding.
For Concept Induction, we find that best results were obtained using  layers 14 to 17, that are the reported results.

\subsection{Hyperparameters}
For \citet{eyal-etal-2022-large}, we tried different resolution, varying it from 1e-3 to 10, for the Louvain clustering but found very little to no effect.

For Kmeans at the local level, we varied the number of clusters $k$ between 2 and 10.
For Agglomerative clustering at both levels, we tried single, average and complete linkage.

The distance threshold in Agglo $\tau$ was indexed on the distribution of distances.
We fixed an hyperparameter $\nu$ and derived $\tau=\operatorname{avg}(d) - \nu.\operatorname{std}(d)$ with $d$ the distribution of distances between clustered instances.
We made $\nu$ vary between -4 and +8.
For global Kmeans, the number of clusters was indexed using a proportion $\pi$ on the number of lemmas (e.g. $120\%\times\vocab$), $\pi$ varying from 40\% to 400\%.
This may help transfering hyperaparameters to other dataset in future research.

Best hyperaparameters choices are in Table \ref{tab:best-hypers}

\section{Concept Clusters Size Distribution}
\label{sec:appdx:size-distrib}

\begin{figure}[h]
    \centering
    \includegraphics[width=\columnwidth]{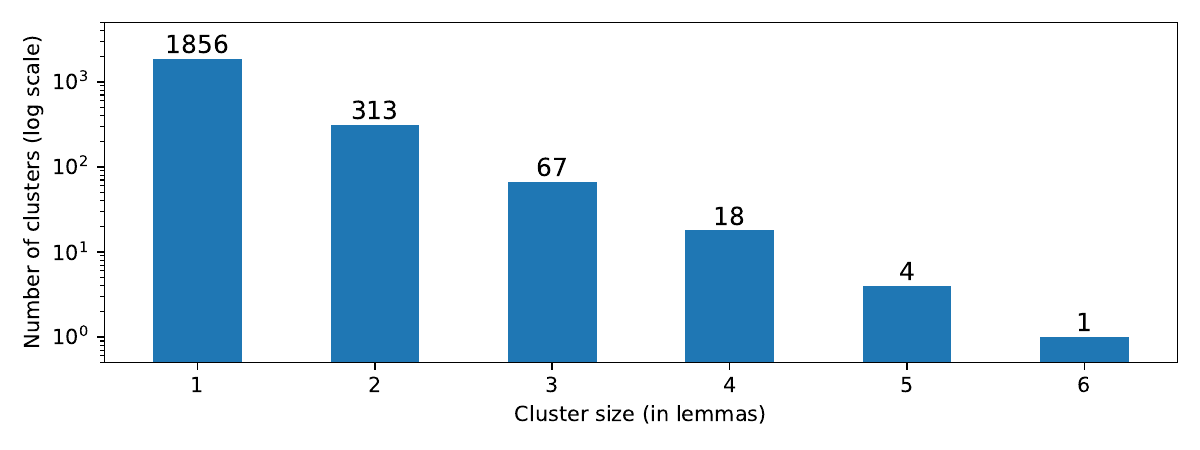}
    \caption{Distribution of cluster size (in number of lemmas) obtained by the Bi-level Agglo system.}
    \label{fig:cluster-size}
\end{figure}

The distribution of the concept cluster size (in number of lemmas) obtained with Bi-level Agglo system can be found in Figure \ref{fig:cluster-size}

\section{Scientific Artifacts}

We used WordNet and SemCor, both properties of Princeton University.
Licence can be found at \url{https://wordnet.princeton.edu/license-and-commercial-use}.

\end{document}